# Using machine learning techniques to predict hospital admission at the emergency department

Georgios Feretzakis[a,b,c †], George Karlis[d, †], Evangelos Loupelis[b], Dimitris Kalles[a], Rea Chatzikyriakou[e], Nikolaos Trakas[f], Eugenia Karakou[f], Aikaterini Sakagianni[g], Lazaros Tzelves[h], Stavroula Petropoulou[b], Aikaterini Tika[i], Ilias Dalainas[i] and Vasileios Kaldis[j]

[a] School of Science and Technology, Hellenic Open University, Patras, Greece

[b] Sismanogleio General Hospital, IT department, Marousi, Greece

[c] Sismanogleio General Hospital, Department of Quality Control, Research and Continuing Education, Marousi, Greece

[d] Thoracic Diseases General Hospital Sotiria, Intensive Care Unit, Athens, Greece

[e] Sismanogleio General Hospital, Hematology Laboratory, Marousi, Greece

[f] Sismanogleio General Hospital, Biochemistry Department, Marousi, Greece

[g] Intensive Care Unit, Sismanogleio General Hospital, 15126 Marousi, Greece

[h] Second Department of Urology, National and Kapodistrian University of Athens, Sismanogleio General Hospital, Athens, Greece

[i] Sismanogleio General Hospital, Administration, Marousi, Greece

[j] Sismanogleio General Hospital, Emergency Department, Marousi, Greece

[†] Equal contribution (1st authors)

Corresponding author: Georgios Feretzakis, email: georgios.feretzakis@ac.eap.gr

**Abstract**

**Introduction**: One of the most important tasks in the Emergency Department (ED) is to promptly identify the patients who will benefit from hospital admission. Machine Learning (ML) techniques show promise as diagnostic aids in healthcare.

**Material and methods**: We investigated the following features seeking to investigate their performance in predicting hospital admission: serum levels of Urea, Creatinine, Lactate Dehydrogenase, Creatine Kinase, C-Reactive Protein, Complete Blood Count with differential, Activated Partial Thromboplastin Time, D-Dimer, International Normalized Ratio, age, gender, triage disposition to ED unit and ambulance utilization. A total of 3,204 ED visits were analyzed.

**Results**: The proposed algorithms generated models which demonstrated acceptable performance in predicting hospital admission of ED patients. The range of F-measure and ROC Area values of all eight evaluated algorithms were [0.679-0.708] and [0.734-0.774], respectively.

**Discussion**: The main advantages of this tool include easy access, availability, yes/no result, and low cost. The clinical implications of our approach might facilitate a shift from traditional clinical decision-making to a more sophisticated model.

**Conclusion**: Developing robust prognostic models with the utilization of common biomarkers is a project that might shape the future of emergency medicine. Our findings warrant confirmation with implementation in pragmatic ED trials.

# 1. Introduction

The Emergency Department (ED) represents a key element of any given healthcare facility and retains a high public profile. ED staff manage patients with a huge variety of medical problems and deal with all sorts of emergencies. ED congestion resulting in delays in care remains a frequent issue that prompts the development of tools for rapid triage of high-risk patients (Sun et al. 2012). Moreover, it is well documented that timely interventions are critical for several acute diseases (Gaieski et al. 2017, Rathore et al. 2009). One of the most commonly encountered ED priorities is to quickly identify those who will need hospital admission. Traditionally, this decision relies on clinical judgment aided by the results of laboratory tests. Human factors leading to diagnostic errors occur frequently and are associated with increased morbidity and mortality (Hautz et al. 2019).

Machine Learning (ML) techniques show promise as diagnostic aids in healthcare and have sparked the discussion for their wider application in the ED (Feretzakis et al. 2020). Developing robust prognostic models with the utilization of common biomarkers to facilitate rapid and reliable decision-making regarding hospital admission of ED patients is a project that might shape the future of emergency medicine. However, relevant data from the ED is scarce. Recent studies have focused on clinical outcome and mortality prediction (Raita et al. 2019, Yan et al. 2020).

We investigated biochemical markers and coagulation tests that are routinely checked in patients visiting the ED, seeking to investigate their performance in predicting whether the patients will be admitted to the hospital. Our aim is to find an algorithm using ML techniques to assist clinical decision-making in the emergency setting.

## 2. Materials and methods

This research is a retrospective observational study conducted in the ED of a public tertiary care hospital in Greece that has been approved by the Institutional Review Board of Sismanogleio General Hospital (Ref.No 15177/2020, 5969/2021).

This study examines the performance of eight machine learning models based on data of the Biochemistry and Hematology Departments from ED patients. Blood samples were obtained for the measurement of biochemical and hematological parameters. The serum levels of Urea (UREA) [Normal Range (NR)=10-50 mg/dL-test principle: kinetic test with urease and glutamate dehydrogenase], Creatinine (CREA) (NR=0.5-1.5 mg/dL-kinetic colorimetric assay based on the Jaffé method), Lactate Dehydrogenase (LDH) (NR=135-225 U/L-UV assay), Creatine Kinase (CPK) (NR=25-190 U/L-UV assay), C-Reactive Protein (CRP) (NR < 6 mg/L-particle-enhanced immunoturbidimetric assay) were measured using the Cobas 6000 c501 Analyzer (Roche Diagnostics, Mannheim, Germany). Complete blood count (CBC) samples were collected, and parameters such as White Blood Cell (WBC) (NR=4-11 K/μl-flow cytometry analysis), Neutrophil (NEUT) (NR=40-75 %-flow cytometry), Lymphocyte (LYM) (NR=20-40%-flow cytometry) and Platelet (PLT) (NR=150-400 K/μl-hydrodynamic focusing-flow cytometry) counts and Hemoglobin (HGB) (NR=12-17.5 g/dL-SLS method) were analyzed using the Sysmex XE 2100 Automated Hematology Analyzer (Sysmex Corporation, Kobe, Japan). Routine hemostasis parameters such as activated partial thromboplastin time (aPTT) (NR=24-39 sec-clotting method), D-Dimer (DD) (NR <500 μg/L-immunoturbidimetric assay), and International Normalized Ratio (INR) (NR=0.86-1.20-calculated) were determined in plasma using the

BCS XP Automated Hemostasis Analyzer (Siemens Healthcare Diagnostics, Marburg, Germany).

All raw data was retrieved from a standard Hospital Information System (HIS) and a Laboratory Information System (LIS). The analysis was performed using the Waikato Environment for Knowledge Analysis (WEKA) (Hall et al. 2009), a Data Mining Software in Java workbench.

A total of 3,204 ED visits were analyzed during the study period (14 March – 4 May 2019). The anonymous data set under investigation contains eighteen features presented in Table 1.

**Table 1.** Features

| Features | Type | Mean | Standard Deviation |
|---|---|---|---|
| CPK | numerical | 179.155 | 1183.877 |
| CREA | numerical | 1.06 | 0.827 |
| CRP | numerical | 39.094 | 71.48 |
| LDH | numerical | 222.327 | 156.343 |
| UREA | numerical | 45.651 | 33.616 |
| aPTT | numerical | 34.227 | 11.443 |
| DDIMER | numerical | 1422.899 | 2522.921 |
| INR | numerical | 1.131 | 0.571 |
| HGB | numerical | 12.87 | 2.13 |
| LYM | numerical | 22.085 | 11.672 |
| NEUT | numerical | 69.478 | 13.083 |
| PLT | numerical | 252.467 | 87.814 |
| WBC | numerical | 9.617 | 5.153 |
| Age | numerical; Integer* | 61.175 | 20.822 |
| Gender | categorical {Male, Female} | | |
| ED Unit | categorical {Urology, Pulmonology, Internal Medicine, Otolaryngology, Triage, Cardiology, General Surgery, Opthalmology, Vascular Surgery, Thoracic Surgery} | | |
| Ambulance | Categorical {Yes, No} | | |
| Admission | Categorical {Yes, No} | | |

*Patients' age has been rounded to the nearest whole number

To assess the performance of the best-performing model (Smith and Frank 2016) for our analysis in WEKA, we have used a 10-fold cross-validation approach to avoid overfitting; Cross-validation is widely regarded as a quite reliable way to assess the quality of results from machine learning techniques. WEKA (Kasperczuk and Dardzińska 2016, Bouckaert 2004, Han et al. 2000) provides detailed results for the classifiers under investigation regarding the following evaluation measures:

a) TP Rate: rate of true positives (instances correctly classified as a given class);
b) FP Rate: rate of false positives (instances falsely classified as a given class);
c) Precision: proportion of instances that are truly of a class divided by the total instances classified as that classes;
d) Recall: proportion of instances classified as a given class divided by the actual total in that class (equivalent to TP rate);
e) F-Measure: a general indicator of the quality of the model;
f) MMC: a correlation coefficient calculated from all four values of the confusion matrix.
g) The area under the Receiver Operating Characteristics (ROC) curve (AUC) (ROC Area): a graphical plot that illustrates the performance of a binary classifier system as its discrimination threshold is varied. The accuracy of the test depends on how well the test separates the group being tested into those with and without the disease in question. Accuracy is measured by the area under the ROC curve;
h) The PRC plot shows the relationship between precision and sensitivity.

Among many algorithms that were evaluated for our research purposes, in this article, we present only the eight best-performing algorithms, mainly in terms of ROC area and F-Measure.

During our experiments, we retained the default settings of all classification algorithms' original implementations provided by WEKA. Each algorithm was evaluated on two data sets; the original data set, including the missing values, and on the data set where the missing values were identified, and they were replaced with appropriate values using WEKA's *ReplaceMissingValues filter*. Furthermore, since the number of patients in our data set who met clinical criteria for hospital admission (36.7%) is less than those who did not meet (63.3%), we applied WEKA's ClassBalancer technique (Hall et al. 2009) to prevent overfitting by reweighting the instances in the data set so that each class had the same total weight during the phase of model training.

In our investigation, we evaluated a Naive Bayes classifier (Rish 2005, John and Langley 1995), a multinomial logistic regression model with a ridge estimator (le Cessie and van Houwelingen 1992), two boosting techniques; AdaBoost (Freund and Schapire 1996) and LogitBoost (Friedman et al. 2020), Classification via Regression (Frank et al. 1998), a random forest (Breiman 2001), a bagging method (Breiman 1996) and a multilayer perceptron (MLP) (a neural network trained with error backpropagation) (Hall et al. 2009, Smith and Frank 2016).

## 3. Results

The performance of each algorithm was evaluated on its ability to predict whether a patient seen in the emergency department is subsequently admitted to the hospital or not by only taking into consideration the features presented in Table 1. All algorithms were evaluated on both datasets (original with missing values, modified by using the *ReplaceMissingValues* filter), and the detailed results are presented in Appendix (Tables A1-A16). The classification performance's results on the original data set, regarding the F-Measure and ROC Area of each algorithm, are summarized in Table 2 and Figure 1.

**Table 2.** Weighted Average values of F-Measure and ROC Area for all methods (10-fold cross-validation)

|  | F-Measure | ROC Area |
|---|---|---|
| NaiveBayes | 0.679 | 0.734 |
| Logistic Regression | 0.697 | 0.762 |
| Ada boost | 0.685 | 0.753 |
| **Logit boost** | **0.708** | **0.774** |
| ClassificationViaRegression | 0.691 | 0.760 |
| Random Forest | 0.689 | 0.757 |
| Bagging | 0.703 | 0.764 |
| Multilayer perceptron | 0.707 | 0.742 |

**Figure 1.** Weighted Average values of F-Measure and ROC Area for all methods (10-fold cross-validation)

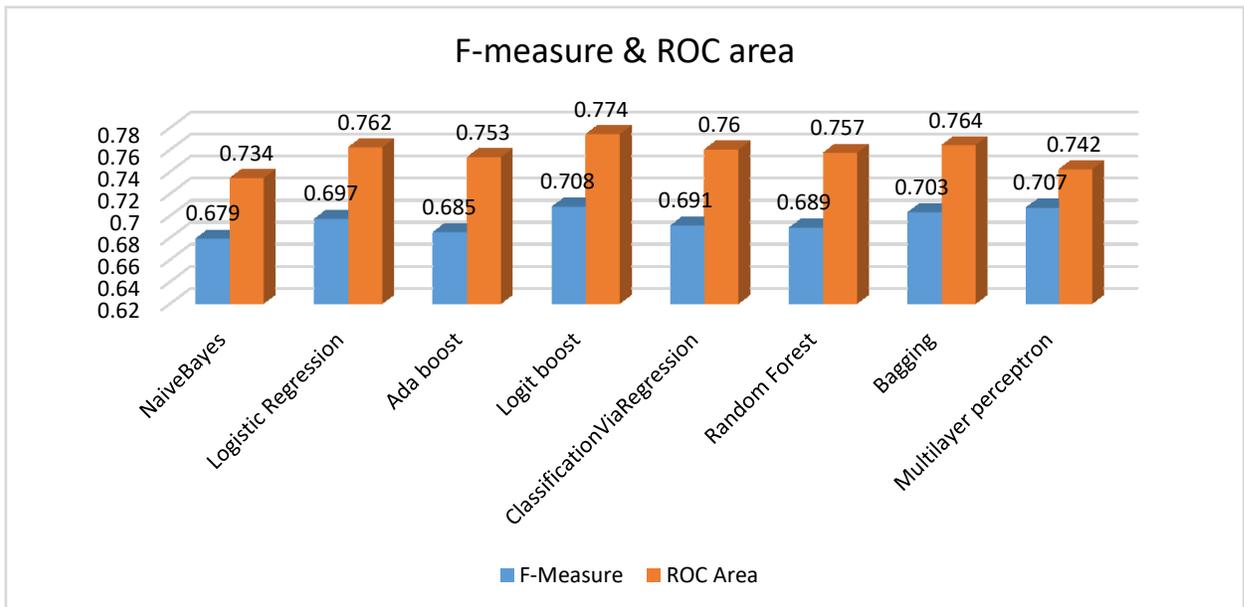

According to Table 2, considering the weighted average values, it can be seen that Logit boost slightly outperformed other models with respect to both F-measure and ROC

Area with values of 0.708 and 0.774, respectively. We can also observe that the range of F-measure and ROC Area values of all eight algorithms that were evaluated are [0.679-0.708] and [0.734-0.774], respectively, and they can be considered acceptable (Mandrekar 2010). The classification performance's results of F-Measure and ROC Area on the data set where the missing values have been replaced by using WEKA's ReplaceMissingValues filter are summarized in Table 3 and Figure 2.

**Table 3.** Weighted Average values of F-Measure and ROC Area for all methods -ReplaceMissingValues filters (10-fold cross-validation)

|  | F-Measure | ROC Area |
|---|---|---|
| NaiveBayes | 0.663 | 0.741 |
| Logistic Regression | 0.696 | 0.765 |
| Ada boost | 0.674 | 0.731 |
| Logit boost | 0.704 | 0.757 |
| ClassificationViaRegression | 0.691 | 0.758 |
| **Random Forest** | **0.723** | **0.789** |
| Bagging | 0.712 | 0.775 |
| Multilayer perceptron | 0.697 | 0.740 |

**Figure 2.** Weighted Average values of F-Measure and ROC Area for all methods - ReplaceMissingValues filters (10-fold cross-validation)

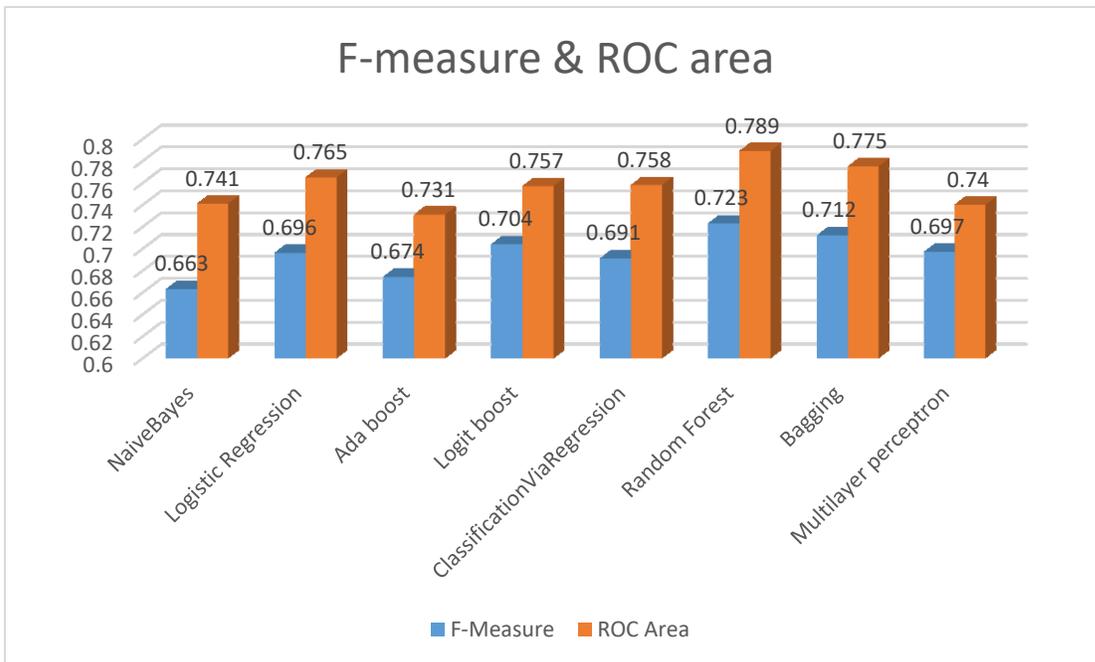

According to Table 3, considering the weighted average values, it can be seen that the Random Forest slightly outperformed other models with respect to both F-measure and ROC Area with values of 0.723 and 0.789, respectively. Additionally, we can observe that the range of F-measure and ROC Area values of all eight algorithms are [0.663-0.723] and [0.731-0.789], respectively, and as previously noted, they can also be considered acceptable. Furthermore, we were positively surprised to see that the impact of missing values on the classifiers' performance was less pronounced than we initially thought.

Furthermore, since the admitted patients were 1175 versus 2029 that were not admitted, we applied the WEKA's ClassBalancer technique on both datasets and re-evaluate the performance of the two classifiers (Logit boost and Random Forest). After the application of the ClassBalancer filter in the original data set, we observe that the performance of Logit boost (F-measure:0.693; ROC area:0.773) (Table A17) is quite similar to this of the imbalanced data set (F-measure:0.708; ROC area:0.774). Similar behavior, we also observe in the performance of Random Forest before (F-measure:0.723; ROC area:0.789) and after (F-measure:0.704; ROC area:0.784) (Table A18) the application of ClassBalancer filter in the data set where the missing values have been replaced by using the ReplaceMissingValues filter.

## 4. Discussion

Based on the data from 3,204 adult ED visits, using common laboratory tests and basic demographics, we evaluated eight ML algorithms that generated models that can reliably predict the hospital admission of patients seen in the ED. Our study utilized pre-existing patient data from a standard HIS and LIS. Therefore, the methods proposed here can serve as a valuable tool for the clinician to decide whether to admit or not an ED patient. The main advantages of this tool include easy access, availability, yes/no result, and low cost. The clinical implications of our approach might be significant and might facilitate a shift from traditional clinical decision-making to a more sophisticated model.

The application of machine learning techniques in the ED is not entirely new. Yet, it is not considered the standard of care. Current efforts are aiming to develop and integrate clinical decision support systems able to provide objective criteria to healthcare professionals. Our study is consistent with previous research showing that logistic regression is the most frequently used technique for model design. The area under the receiver operating curve (AUC) is the most frequently used performance measure (Fernandes et al. 2020). Moreover, the major goal of such predictive tools is to identify high-risk patients accurately and differentiate them from stable, low-risk patients that can be safely discharged from the ED (Levin et al. 2017) and communicate this identification to the medical expert who can take this information into account while making a decision on admission or discharge.

The hectic pace of work and the stressful setting of the ED have negative consequences on patient safety (Weant et al 2014, Morley et al. 2018). It is well established that human factors play an important role in the efficiency of healthcare systems. Different error types have different underlying mechanisms and require specific methods of risk management (Reason 1995). A fearful shortcoming for the emergency physician is to fail to admit a seriously ill patient. Our methods might be useful to reduce these errors while explicitly acknowledging that they are meant to aid and not substitute clinical judgment.

This study is not without limitations. In our analysis, we did not include clinical parameters such as the vital signs and the Emergency Severity Index (ESI) (González and Soltero 2009). There were also missing values in the data we collected and analyzed; for example, not all of the analyzed ED visits had all the laboratory investigations available. Furthermore, our preliminary findings have not yet been followed up by an implementation phase, and the proposed algorithms have not been validated in a pragmatic ED trial. Therefore, future research is warranted in order to demonstrate whether they can actually improve care.

In summary, we present an inexpensive clinical decision support tool derived from readily available patient data. This tool is intended to aid the emergency physician regarding hospital admission decisions, as the development of machine learning models represents a rapidly evolving field in healthcare.

**5. Conclusions**

In this study, we evaluated a collection of very popular ML classifiers on data from an ED. The proposed algorithms generated models which demonstrated acceptable performance in predicting hospital admission of ED patients based on common biochemical markers, coagulation tests, basic demographics, ambulance utilization, and triage disposition to the ED unit. Our research confirms the prevalent current notion that the utilization of artificial intelligence may have a favorable impact on the future of emergency medicine.

**Clinical significance**

- The Emergency Department (ED) represents a key element of any given healthcare facility and retains a high public profile.

- Machine Learning (ML) techniques show promise as diagnostic aids in healthcare and have sparked the discussion for their wider application in the ED.

- Based on the data from 3,204 adult ED visits, using common laboratory tests and basic demographics, we evaluated eight ML algorithms that generated models that can reliably predict the hospital admission of patients seen in the ED.

- We present an inexpensive clinical decision support tool derived from readily available patient data. This tool is intended to aid the emergency physician regarding hospital admission decisions.

**Disclosure statement**

The authors report no conflict of interest.

# Appendix

**Table A1.** Performance results by class of *NaiveBayes* (10-fold cross-validation)

|  | TP Rate | FP Rate | Precision | Recall | F-Measure | MCC | ROC Area | PRC Area | Admission |
|---|---|---|---|---|---|---|---|---|---|
|  | 0.360 | 0.091 | 0.696 | 0.360 | 0.474 | 0.330 | 0.734 | 0.602 | Yes |
|  | 0.909 | 0.640 | 0.710 | 0.909 | 0.797 | 0.330 | 0.734 | 0.811 | No |
| Weighted Avg. | 0.708 | 0.439 | 0.705 | 0.708 | 0.679 | 0.330 | 0. 734 | 0.734 |  |

**Table A2.** Performance results by class of *NaiveBayes* –ReplaceMissingValues filters (10-fold cross-validation)

|  | TP Rate | FP Rate | Precision | Recall | F-Measure | MCC | ROC Area | PRC Area | Admission |
|---|---|---|---|---|---|---|---|---|---|
|  | 0.329 | 0.091 | 0.677 | 0.329 | 0.442 | 0.300 | 0.741 | 0.603 | Yes |
|  | 0.909 | 0.671 | 0.700 | 0.909 | 0.791 | 0.300 | 0.741 | 0.822 | No |
| Weighted Avg. | 0.696 | 0.458 | 0.692 | 0.696 | 0.663 | 0.300 | 0.741 | 0.742 |  |

**Table A3.** Performance results by class of Logistic Regression (10-fold cross-validation)

|  | TP Rate | FP Rate | Precision | Recall | F-Measure | MCC | ROC Area | PRC Area | Admission |
|---|---|---|---|---|---|---|---|---|---|
|  | 0.456 | 0.142 | 0.650 | 0.456 | 0.536 | 0.346 | 0.762 | 0.641 | Yes |
|  | 0.858 | 0.544 | 0.732 | 0.858 | 0.790 | 0.346 | 0.762 | 0.838 | No |
| Weighted Avg. | 0.711 | 0.396 | 0.702 | 0.711 | 0.697 | 0.346 | 0.762 | 0.766 |  |

**Table A4.** Performance results by class of Logistic Regression–ReplaceMissingValues filters (10-fold cross-validation)

|  | TP Rate | FP Rate | Precision | Recall | F-Measure | MCC | ROC Area | PRC Area | Admission |
|---|---|---|---|---|---|---|---|---|---|
|  | 0.456 | 0.143 | 0.649 | 0.456 | 0.536 | 0.345 | 0.765 | 0.643 | Yes |
|  | 0.857 | 0.544 | 0.731 | 0.857 | 0.789 | 0.345 | 0.765 | 0.841 | No |
| Weighted Avg. | 0.710 | 0.397 | 0.701 | 0.710 | 0.696 | 0.345 | 0.765 | 0.768 |  |

**Table A5.** Performance results by class of AdaBoost (10-fold cross-validation)

|  | TP Rate | FP Rate | Precision | Recall | F-Measure | MCC | ROC Area | PRC Area | Admision |
|---|---|---|---|---|---|---|---|---|---|
|  | 0.423 | 0.137 | 0.642 | 0.423 | 0.510 | 0.323 | 0.753 | 0.620 | Yes |
|  | 0.863 | 0.577 | 0.721 | 0.863 | 0.786 | 0.323 | 0.753 | 0.836 | No |
| Weighted Avg. | 0.702 | 0.415 | 0.692 | 0.702 | 0.685 | 0.323 | 0.753 | 0.757 |  |

**Table A6.** Performance results by class of AdaBoost–ReplaceMissingValues filters (10-fold cross-validation)

|  | TP Rate | FP Rate | Precision | Recall | F-Measure | MCC | ROC Area | PRC Area | Admission |
|---|---|---|---|---|---|---|---|---|---|
|  | 0.396 | 0.133 | 0.634 | 0.396 | 0.487 | 0.302 | 0.731 | 0.604 | Yes |
|  | 0.867 | 0.604 | 0.713 | 0.867 | 0.782 | 0.302 | 0.731 | 0.815 | No |
| Weighted Avg. | 0.694 | 0.431 | 0.684 | 0.694 | 0.674 | 0.302 | 0.731 | 0.738 |  |

**Table A7.** Performance results by class of LogitBoost (10-fold cross-validation)

|  | TP Rate | FP Rate | Precision | Recall | F-Measure | MCC | ROC Area | PRC Area | Admission |
|---|---|---|---|---|---|---|---|---|---|
|  | 0.489 | 0.147 | 0.658 | 0.489 | 0.561 | 0.370 | 0.774 | 0.657 | Yes |
|  | 0.853 | 0.511 | 0.742 | 0.853 | 0.794 | 0.370 | 0.774 | 0.854 | No |
| Weighted Avg. | 0.719 | 0.377 | 0.711 | 0.719 | 0.708 | 0.370 | 0.774 | 0.782 |  |

**Table A8.** Performance results by class of LogitBoost –ReplaceMissingValues filters (10-fold cross-validation)

|  | TP Rate | FP Rate | Precision | Recall | F-Measure | MCC | ROC Area | PRC Area | Admission |
|---|---|---|---|---|---|---|---|---|---|
|  | 0.464 | 0.135 | 0.666 | 0.464 | 0.547 | 0.364 | 0.757 | 0.641 | Yes |
|  | 0.865 | 0.536 | 0.736 | 0.865 | 0.795 | 0.364 | 0.757 | 0.837 | No |
| Weighted Avg. | 0.718 | 0.389 | 0.710 | 0.718 | 0.704 | 0.364 | 0.757 | 0.765 |  |

**Table A9.** Performance results by class of ClassificationViaRegression (10-fold cross-validation)

|  | TP Rate | FP Rate | Precision | Recall | F-Measure | MCC | ROC Area | PRC Area | Admission |
|---|---|---|---|---|---|---|---|---|---|
|  | 0.447 | 0.145 | 0.641 | 0.447 | 0.527 | 0.334 | 0.760 | 0.639 | Yes |
|  | 0.855 | 0.553 | 0.727 | 0.855 | 0.786 | 0.334 | 0.760 | 0.839 | No |
| Weighted Avg. | 0.705 | 0.403 | 0.696 | 0.705 | 0.691 | 0.334 | 0.760 | 0.766 |  |

**Table A10.** Performance results by class of ClassificationViaRegression–ReplaceMissingValues filters (10-fold cross-validation)

|  | TP Rate | FP Rate | Precision | Recall | F-Measure | MCC | ROC Area | PRC Area | Admission |
|---|---|---|---|---|---|---|---|---|---|
|  | 0.447 | 0.145 | 0.641 | 0.447 | 0.527 | 0.334 | 0.758 | 0.638 | Yes |
|  | 0.855 | 0.553 | 0.727 | 0.855 | 0.786 | 0.334 | 0.758 | 0.837 | No |

|  | TP Rate | FP Rate | Precision | Recall | F-Measure | MCC | ROC Area | PRC Area |
|---|---|---|---|---|---|---|---|---|
| Weighted Avg. | 0.705 | 0.403 | 0.696 | 0.705 | 0.691 | 0.334 | 0.758 | 0.764 |

**Table A11.** Performance results by class of Random Forest (10-fold cross-validation)

|  | TP Rate | FP Rate | Precision | Recall | F-Measure | MCC | ROC Area | PRC Area | Admission |
|---|---|---|---|---|---|---|---|---|---|
|  | 0.394 | 0.103 | 0.689 | 0.394 | 0.501 | 0.345 | 0.757 | 0.650 | Yes |
|  | 0.897 | 0.606 | 0.719 | 0.897 | 0.798 | 0.345 | 0.757 | 0.832 | No |
| Weighted Avg. | 0.713 | 0.422 | 0.708 | 0.713 | 0.689 | 0.345 | 0.757 | 0.765 |  |

**Table A12.** Performance results by class of Random Forest–ReplaceMissingValues filters (10-fold cross-validation)

|  | TP Rate | FP Rate | Precision | Recall | F-Measure | MCC | ROC Area | PRC Area | Admission |
|---|---|---|---|---|---|---|---|---|---|
|  | 0.540 | 0.161 | 0.660 | 0.540 | 0.594 | 0.399 | 0.789 | 0.676 | Yes |
|  | 0.839 | 0.460 | 0.759 | 0.839 | 0.797 | 0.399 | 0.789 | 0.858 | No |
| Weighted Avg. | 0.729 | 0.350 | 0.723 | 0.729 | 0.723 | 0.399 | 0.789 | 0.791 |  |

**Table A13.** Performance results by class of Bagging (10-fold cross-validation)

|  | TP Rate | FP Rate | Precision | Recall | F-Measure | MCC | ROC Area | PRC Area | Admission |
|---|---|---|---|---|---|---|---|---|---|
|  | 0.471 | 0.142 | 0.657 | 0.471 | 0.549 | 0.360 | 0.764 | 0.654 | Yes |
|  | 0.858 | 0.529 | 0.737 | 0.858 | 0.793 | 0.360 | 0.764 | 0.840 | No |
| Weighted Avg. | 0.716 | 0.387 | 0.708 | 0.716 | 0.703 | 0.360 | 0.764 | 0.772 |  |

**Table A14.** Performance results by class of Bagging–ReplaceMissingValues filters (10-fold cross-validation)

|  | TP Rate | FP Rate | Precision | Recall | F-Measure | MCC | ROC Area | PRC Area | Admission |
|---|---|---|---|---|---|---|---|---|---|
|  | 0.515 | 0.161 | 0.649 | 0.515 | 0.574 | 0.375 | 0.775 | 0.654 | Yes |
|  | 0.839 | 0.485 | 0.749 | 0.839 | 0.791 | 0.375 | 0.775 | 0.852 | No |
| Weighted Avg. | 0.720 | 0.366 | 0.712 | 0.720 | 0.712 | 0.375 | 0.775 | 0.779 |  |

**Table A15.** Performance results by class of Multilayer perceptron (10-fold cross-validation)

|  | TP Rate | FP Rate | Precision | Recall | F-Measure | MCC | ROC Area | PRC Area | Admission |
|---|---|---|---|---|---|---|---|---|---|

|  | TP Rate | FP Rate | Precision | Recall | F-Measure | MCC | ROC Area | PRC Area | Admission |
|---|---|---|---|---|---|---|---|---|---|
|  | 0.542 | 0.190 | 0.623 | 0.542 | 0.580 | 0.364 | 0.742 | 0.622 | Yes |
|  | 0.810 | 0.458 | 0.753 | 0.810 | 0.781 | 0.364 | 0.742 | 0.815 | No |
| Weighted Avg. | 0.712 | 0.360 | 0.705 | 0.712 | 0.707 | 0.364 | 0.742 | 0.744 |  |

**Table A16.** Performance results by class of Multilayer perceptron–ReplaceMissingValues filters (10-fold cross-validation)

|  | TP Rate | FP Rate | Precision | Recall | F-Measure | MCC | ROC Area | PRC Area | Admission |
|---|---|---|---|---|---|---|---|---|---|
|  | 0.480 | 0.161 | 0.633 | 0.480 | 0.546 | 0.343 | 0.740 | 0.617 | Yes |
|  | 0.839 | 0.520 | 0.736 | 0.839 | 0.784 | 0.343 | 0.740 | 0.808 | No |
| Weighted Avg. | 0.707 | 0.388 | 0.698 | 0.707 | 0.697 | 0.343 | 0.740 | 0.738 |  |

**Table A17.** Performance results by class of Logit Boost– ClassBalancer filter (10-fold cross-validation)

|  | TP Rate | FP Rate | Precision | Recall | F-Measure | MCC | ROC Area | PRC Area | Admission |
|---|---|---|---|---|---|---|---|---|---|
|  | 0.685 | 0.300 | 0.696 | 0.685 | 0.690 | 0.385 | 0.773 | 0.758 | Yes |
|  | 0.700 | 0.315 | 0.690 | 0.700 | 0.695 | 0.385 | 0.773 | 0.779 | No |
| Weighted Avg. | 0.693 | 0.307 | 0.693 | 0.693 | 0.693 | 0.385 | 0.773 | 0.769 |  |

**Table A18.** Performance results by class of Random Forest– ReplaceMissingValues and ClassBalancer filters (10-fold cross-validation)

|  | TP Rate | FP Rate | Precision | Recall | F-Measure | MCC | ROC Area | PRC Area | Admission |
|---|---|---|---|---|---|---|---|---|---|
|  | 0.653 | 0.243 | 0.729 | 0.653 | 0.689 | 0.412 | 0.784 | 0.767 | Yes |
|  | 0.757 | 0.347 | 0.686 | 0.757 | 0.720 | 0.412 | 0.784 | 0.783 | No |
| Weighted Avg. | 0.705 | 0.295 | 0.707 | 0.705 | 0.704 | 0.412 | 0.784 | 0.775 |  |